**Advancing Pavement Distress Detection in Developing Countries: A Novel Deep Learning Approach with Locally-Collected Datasets**

**Blessing Agyei Kyem**
PhD Student
North Dakota State University
Department of Civil, Construction and Environmental Engineering
Email: blessing.agyeikyem@ndsu.edu

**Eugene Kofi Okrah Denteh**
Research Assistant
Kwame Nkrumah University of Science and Technology
Department of Civil Engineering
Email: ekdenteh@st.knust.edu.gh

**Joshua Kofi Asamoah**
PhD Student
North Dakota State University
Department of Civil, Construction and Environmental Engineering
Email: joshua.asamoah@ndsu.edu

**Kenneth Adomako Tutu**
Lecturer
Kwame Nkrumah University of Science and Technology
Department of Civil Engineering
Email: kenneth.tutu@knust.edu.gh

**Armstrong Aboah, Corresponding Author**
Assistant Professor
North Dakota State University
Department of Civil, Construction and Environmental Engineering
Email: armstrong.aboah@ndsu.edu

**ABSTRACT**

Road infrastructure maintenance in developing countries faces unique challenges due to resource constraints and diverse environmental factors. This study addresses the critical need for efficient, accurate, and locally-relevant pavement distress detection methods in these regions. We present a novel deep learning approach combining YOLO (You Only Look Once) object detection models with a Convolutional Block Attention Module (CBAM) to simultaneously detect and classify multiple pavement distress types. The model demonstrates robust performance in detecting and classifying potholes, longitudinal cracks, alligator cracks, and raveling, with confidence scores ranging from 0.46 to 0.93. While some misclassifications occur in complex scenarios, these provide insights into unique challenges of pavement assessment in developing countries. Additionally, we developed a web-based application for real-time distress detection from images and videos. This research advances automated pavement distress detection and provides a tailored solution for developing countries, potentially improving road safety, optimizing maintenance strategies, and contributing to sustainable transportation infrastructure development.

**Keywords:** Pavement distress detection, Deep learning, YOLOv5, Convolutional Block Attention Module, Developing countries, Computer vision, Road infrastructure maintenance, Automated inspection, Pavement dataset, Real-time detection

## INTRODUCTION

Road infrastructure plays a significant role in the economic development and social welfare of any nation, particularly in developing countries. Well-maintained road networks facilitate the efficient movement of goods and people, fostering trade and enabling access to essential services (*1*). However, the deterioration of pavements due to various distresses, such as cracks, potholes, and surface defects, poses significant safety risks to motorists and pedestrians alike. These distresses can lead to vehicle damage, increased fuel consumption, and potentially severe accidents, underscoring the importance of promptly detecting and addressing them (2). Developing countries often face daunting challenges in maintaining their road networks due to limited financial resources, technical expertise, and infrastructure constraints (3, 4). Consequently, cost-effective, and efficient methods for detecting and monitoring pavement distress are critically needed to ensure road safety and optimize the allocation of limited resources for maintenance and rehabilitation efforts.

Given the critical need for efficient pavement distress detection methods, it is essential to understand the various types of pavement distresses and their impact on road safety and transportation efficiency, as well as the limitations of traditional detection approaches. In developing countries, traditional methods of detecting these distresses, which often rely on manual visual inspections, are time-consuming, labor-intensive, and prone to human error and subjectivity (*5*). Furthermore, these methods are inherently reactive, as they only address distress after they have already formed and potentially caused significant damage. This results in accelerated pavement degradation, shortened road lifespans, and higher long-term costs—a burden that most developing countries can hardly afford. To shift from reactive maintenance strategies which are normally a burden for developing countries to proactive interventions which are more effective, there is a pressing need for more advanced and automated techniques to detect pavement distress accurately and efficiently at an early stage (*6*). These automated techniques enable predictive maintenance which can disrupt the vicious cycle of pavement degradation, facilitating timely and targeted maintenance strategies in developing countries.

In recent years, the advent of computer vision techniques offers a promising solution to automating the process of pavement distress detection, addressing the limitations of traditional methods (7–11). These techniques improve pavement maintenance practices by enabling early detection and quantification of distress, facilitating timely interventions, and minimizing the need for costly rehabilitation measures (*12*). However, the existing advanced models for pavement distress detections were trained on Western roads and datasets, making it difficult to transfer them to developing countries, primarily for one reason. Thus, these western datasets may not accurately capture the diverse environmental factors, construction materials, and infrastructural challenges prevalent in developing countries. For instance, Figure 1 shows examples of the pavement surface condition from images captured on roads in a developing country that differ in appearance from those captured in Western countries, like the USA.  Consequently, these models often struggle to adapt to the specific contexts of developing countries, resulting in suboptimal performance, and limited practical utility (*13*). In addition, there are no known publicly available pavement distress datasets for developing countries, making it difficult to train models to meet the specific demands. It is therefore crucial to create datasets tailored to the local conditions of developing countries, accounting for factors such as varying climatic conditions, construction practices, and resource constraints. By leveraging locally relevant data and incorporating region-specific characteristics, these models can deliver more accurate and reliable results, enabling effective pavement distress detection and facilitating informed decision-making for infrastructure management in resource-constrained environments.

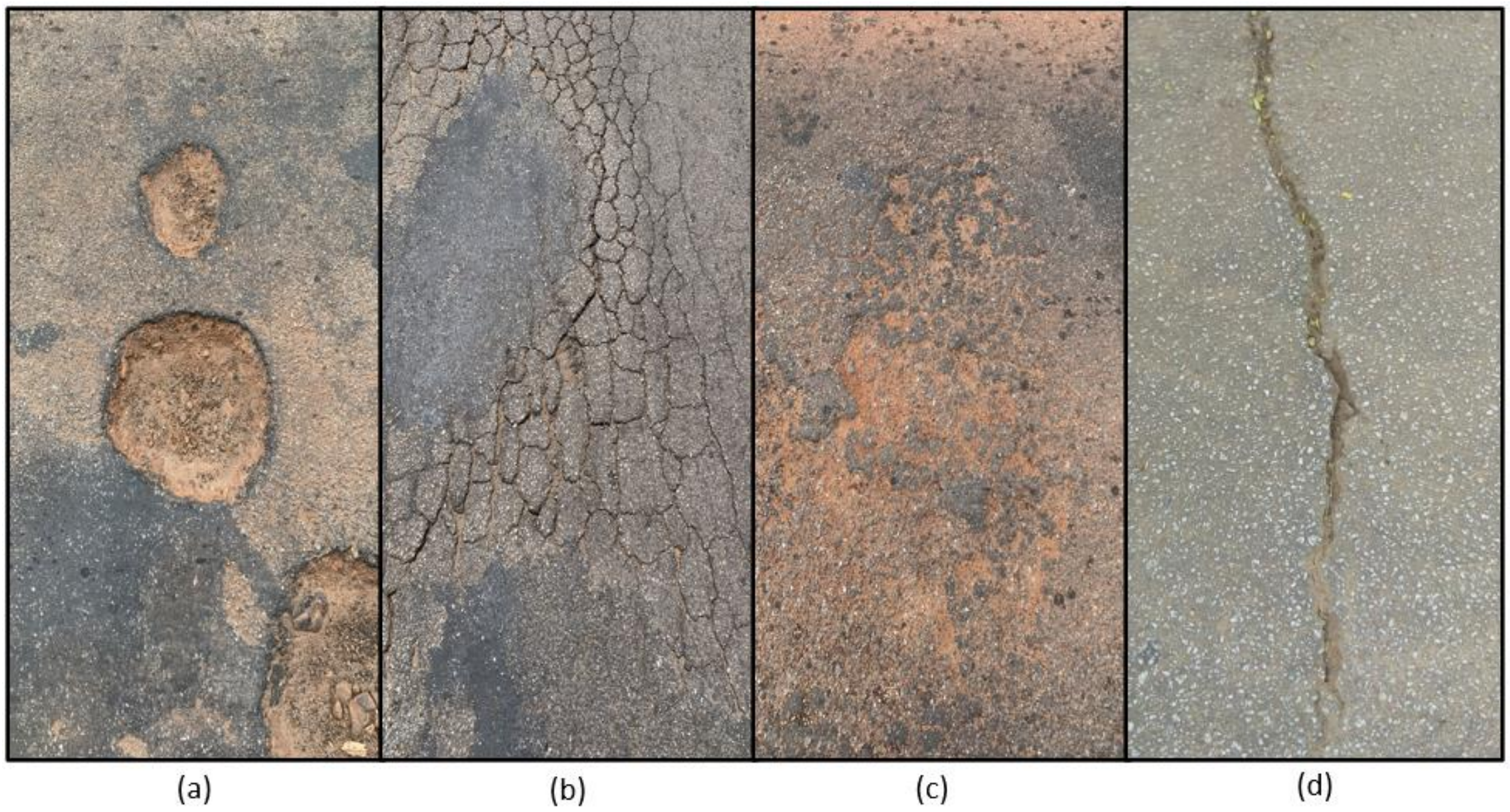

**Figure 1. Samples of pavement images depicting the conditions of roads in a developing country called Ghana**. (a) Potholes (b) Alligator cracks (c) Ravelling (d) Longitudinal crack

Recognizing the limitations of existing computer vision models and the lack of publicly available datasets in developing countries, our research develops a computer vision model and creates a well-annotated publicly available dataset tailored to the roads of developing countries. The study developed a computer vision model that leverages state-of-the-art YOLO (You Only Look Once) object detection *(*14–18) models to accurately detect and classify distresses simultaneously. A crucial aspect of our work is the emphasis on locally relevant data. Since data collection and annotation of pavement distresses is very laborious in developing countries, we have undertaken a comprehensive data collection effort, meticulously capturing visual data from road surfaces in a developing country. This locally sourced dataset serves as the foundation for training and validating our model, ensuring it accurately represents the unique characteristics and challenges encountered in developing countries. By incorporating locally collected data and employing innovative object detection algorithms, our developed model will help detect and classify a wide range of pavement distresses, including fatigue cracking, alligator cracking, raveling, and potholes, in a single computational pipeline. This integrated approach is particularly advantageous in resource-constrained environments, where efficient and comprehensive pavement assessments are critical for optimizing maintenance efforts and maximizing the impact of limited resources. Building upon this innovative model, our research aims to provide a more effective and tailored solution for pavement distress detection in developing countries. By leveraging the power of computer vision and locally relevant data, our work can potentially improve road infrastructure management practices, enabling timely interventions, enhancing road safety, and contributing to the sustainable development of transportation networks in these regions. As part of our efforts, we have made the dataset publicly available for the research community for further research contributions. In addition, we have also developed a web application that leverages this model to detect and classify distress on pavement surfaces from images and videos captured in developing countries. To that end, our research makes the following significant contributions:

1. Curated a well-annotated publicly available pavement dataset tailored to the pavement challenges of developing countries. The dataset consists of bounding box annotations of different pavement distress types. The aim is to facilitate further research in automated pavement inspection methods in developing countries.

2. Proposed an improved state-of-the-art YOLO object detection architecture by integrating the Convolutional Block Attention Module (CBAM), specifically tailoring it for pavement distress detection. This modification enhanced the model's ability to identify subtle distresses present in developing countries.

3. Designed a web-based platform for real-time detection and classification of pavement distresses, optimized for use in developing countries' infrastructure management systems (https://huggingface.co/spaces/Blessing/Asphalt-Pavement-Distresses-Detector).

## RELATED WORK

**Early Pavement Inspection Methods**. Early research in pavement distress detection focused on developing systematic manual inspection methods (19, 20). Wang (21) proposed a standardized visual survey protocol for pavement inspection, which improved consistency but remained time-consuming and subjective. Addressing these limitations, Garbowski et al. (22) introduced one of the first semi-automated systems, combining digital image capture with basic computer analysis for pavement inspection. Their method reduced inspection time compared to fully manual approaches, but accuracy remained a challenge. Building on this, Dhafari et al. (23) developed an image processing algorithm for crack detection, achieving 85% accuracy in controlled settings. However, Yu et al. (24) noted that such algorithms often faltered under varied lighting and road surface conditions, highlighting the need for more robust solutions. These early studies laid the groundwork for automation in pavement distress detection, while also revealing the limitations of simple image processing techniques when applied to the complex task of road surface analysis. With these limitations, there was a need for more robust automated techniques, such as machine learning and deep learning for efficient and accurate pavement distress detection.

**Machine learning and Deep learning approaches**. Building on the limitations of early image processing techniques, researchers began exploring the use of machine learning and deep learning approaches. Sari et al. (25) pioneered the use of Support Vector Machines (SVM) for pavement crack detection, achieving an accuracy of 96.25% on a limited dataset. This marked a significant improvement over previous approaches of image processing. Expanding on this work, Li et al (26) used Random Forests to classify multiple types of pavement distress, demonstrating the potential of machine learning for comprehensive road surface analysis. Despite their improvements, SVM and Random Forest methods had limitations. SVMs struggled with multi-class problems and required careful feature engineering. (27) Random Forests on the other hand, needed large datasets and were computationally expensive for real-time use. Both methods lacked the ability to learn hierarchical features automatically, limiting their effectiveness on complex pavement textures and varying lighting conditions. The advent of deep learning further revolutionized the field. Zhang et al (28) employed Convolutional Neural Networks (CNNs) for crack detection, surpassing the performance of traditional machine learning methods with an accuracy of 97%. Building on CNN architecture, Song et al (29) utilized Faster R-CNN for real-time detection of multiple distress types, addressing the need for efficient, large-scale inspections. More recently, Majidifard et al (30) adapted the YOLO (You Only Look Once) algorithm for pavement distress detection, achieving both high accuracy and real-time performance (5, 8, 31–35). These advancements in machine learning and deep learning techniques have significantly improved the accuracy, speed, and versatility of automated pavement distress detection systems, addressing many of the challenges faced by earlier methods. However, these machine learning and deep learning techniques involve training models on datasets that are only representative of the characteristics of datasets of already developed countries. This makes it extremely hard for these models to generalize well to datasets of developing countries.

**Challenges in Developing Countries**. Despite these significant advancements in automated pavement distress detection, applying these methods in developing countries presents unique challenges. For example, models trained on data from developed nations often perform poorly when applied to roads in developing countries due to differences in construction materials and practices. Additionally, the classification of distress in these developing countries exhibits subtle yet significant differences compared to developed countries due to variations in climate, traffic patterns, and maintenance practices. These observations highlight the need for locally relevant datasets. However, as pointed out by Ojenge (36), there is a scarcity of such datasets from developing regions, especially in Africa. This lack of appropriate training data limits the accuracy and applicability of existing models. Furthermore, various studies have emphasized that resource constraints in developing countries often preclude the use of advanced equipment required for some detection methods. To address these issues, recent studies have explored cost-effective solutions. For instance, Maeda et al (37) proposed using smartphone cameras for data collection of road distress images, while Astor et al (38) investigated the potential of drone surveys for large-scale pavement inspections. These approaches show promise for overcoming resource limitations. Nevertheless, there remains a critical need for pavement distress detection models tailored to developing countries' contexts, considering their unique road conditions, environmental factors, and resource constraints.

## Pavement Distress Dataset

### *Data Collection*

The distress data was collected in a developing country called Ghana. A comprehensive survey of asphalt roads, totaling 51 km, was conducted for this study, encompassing two distinct road networks. The first network, illustrated in Figure 2(a) spans approximately 46.5 kilometers and consists of a diverse range of road environments, including highway and urban sections. This network includes the N6 Highway, a major transportation artery connecting Ejisu to Anloga Junction, providing insights into distress patterns under heavy traffic loads. A section of inner roads in Kumasi, traversing key points like Suame Roundabout and Sofoline Interchange, offered a contrasting perspective on distress caused by frequent stopping, starting, and turning maneuvers. The inclusion of roads in Mampong, Offinso Magazine, and New Bekwai further broadened the study's scope, encompassing roads with potentially different construction histories and maintenance schedules. The second road network, illustrated in Figure 2(b) is situated entirely within the KNUST campus and spans approximately 4.5 kilometers. This network, features roads such as P.V. Obeng Avenue and Ahemfo Avenue, provided data on pavement conditions within a more controlled environment, characterized by potentially lighter traffic and distinct usage patterns. Figure 3 shows the proportion of data collected from the respective locations.

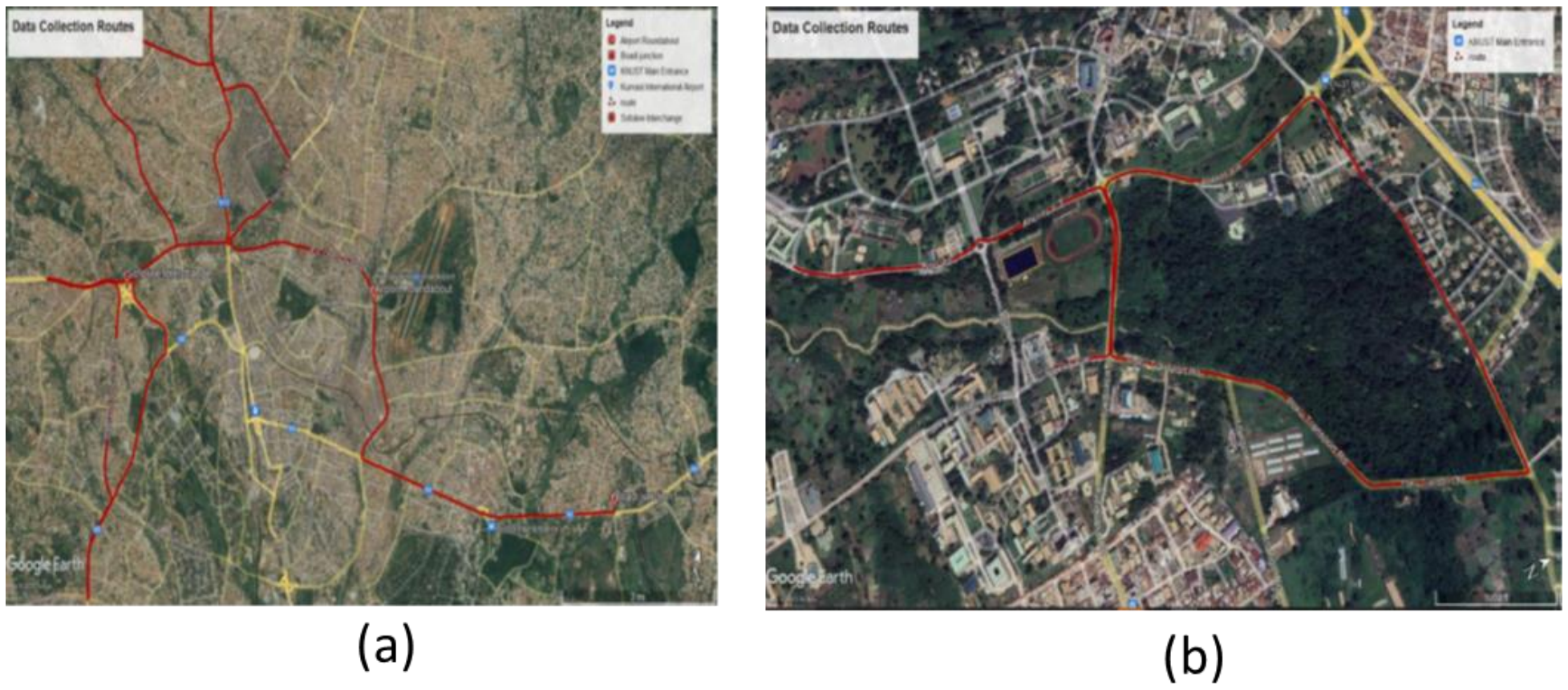

**Figure 2. Selected asphalt road for distress data collection** (a) Asphalt roadways in Kumasi. (b) Asphalt roadways on KNUST Campus

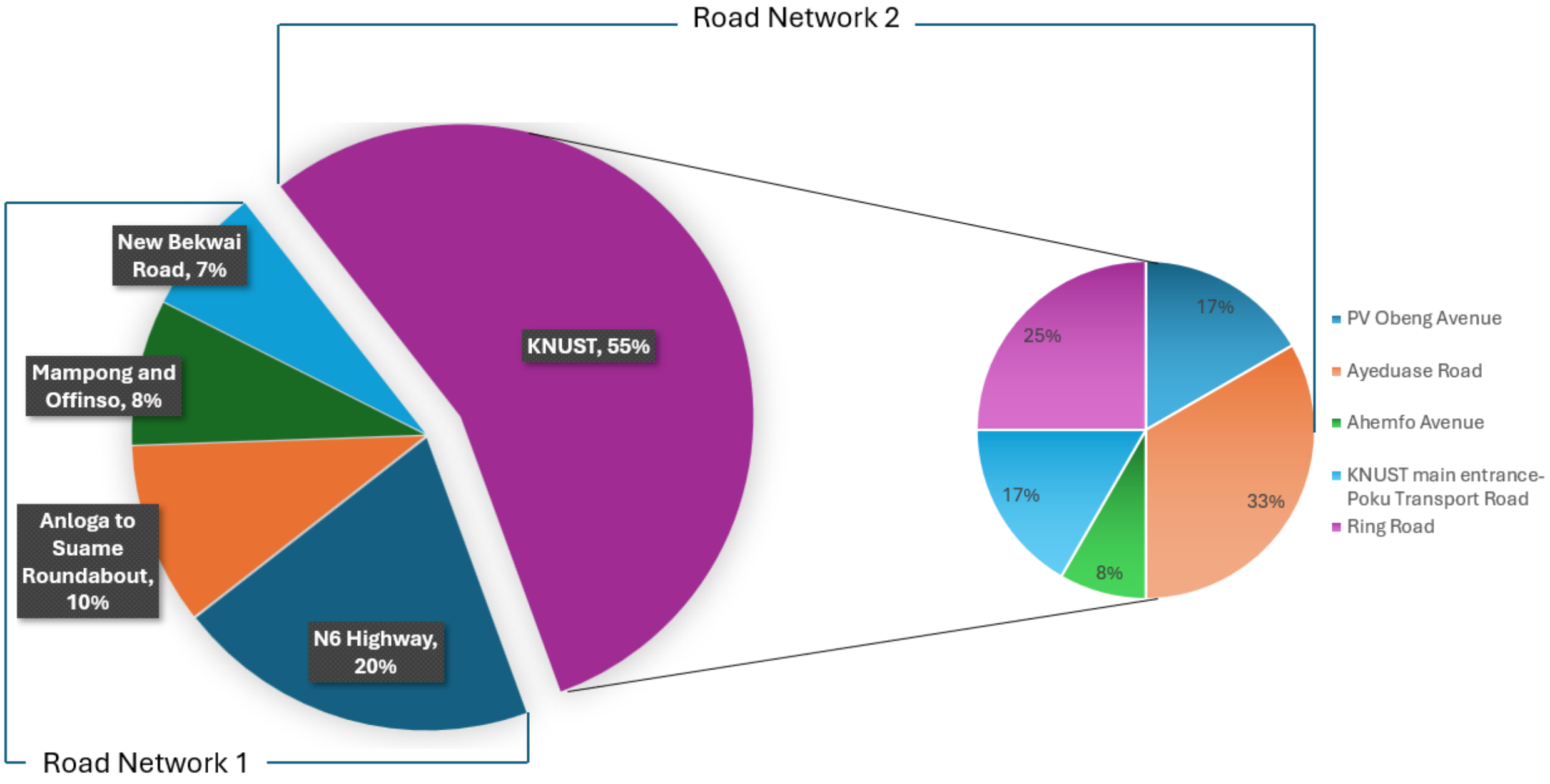

**Figure 3. Pie charts showing the various areas the data was collected from and their proportions**

Different distress data collection approaches were explored in this study. These approaches involved the use of dashboard camera, Google Street view imagery, and smartphones.

Dashcam technology was initially selected for data collection due to its continuous image capture capability and safety benefits. As illustrated in Figure 4, the camera was mounted on the windshield to capture the pavement surface, with the vehicle maintaining an average speed of 15.5 km/hr. While the dashcam's fixed position limited its field of view, particularly near road edges, and produced low-quality images, these limitations ultimately enhanced the robustness and adaptability of the dataset and resulting model. The inclusion of such data provided valuable insights for the model to learn and make predictions on obscured road edge distress images, creating a more robust model capable of handling real-world scenarios with varying image quality and limited visibility. Thus, these limitations inadvertently contributed to a more comprehensive and robust dataset, ultimately leading to an effective pavement distress detection model. Figure 5 showcases examples of these pavement images. Additionally, the dataset's distribution had a natural tendency towards more common distress types, prompting us to seek balanced representation.

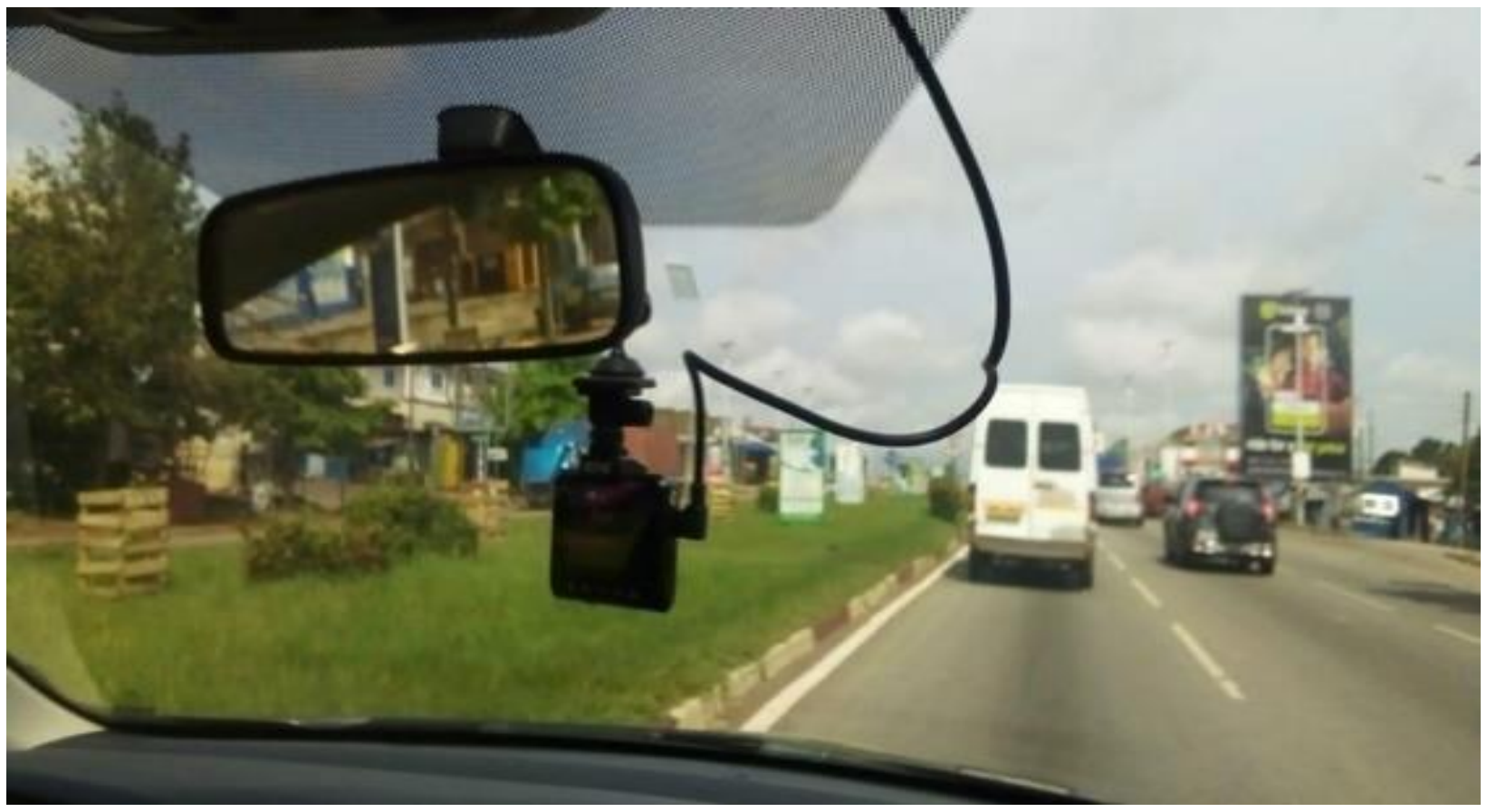

**Figure 4. Front-view of the dashcam used for the data collection**

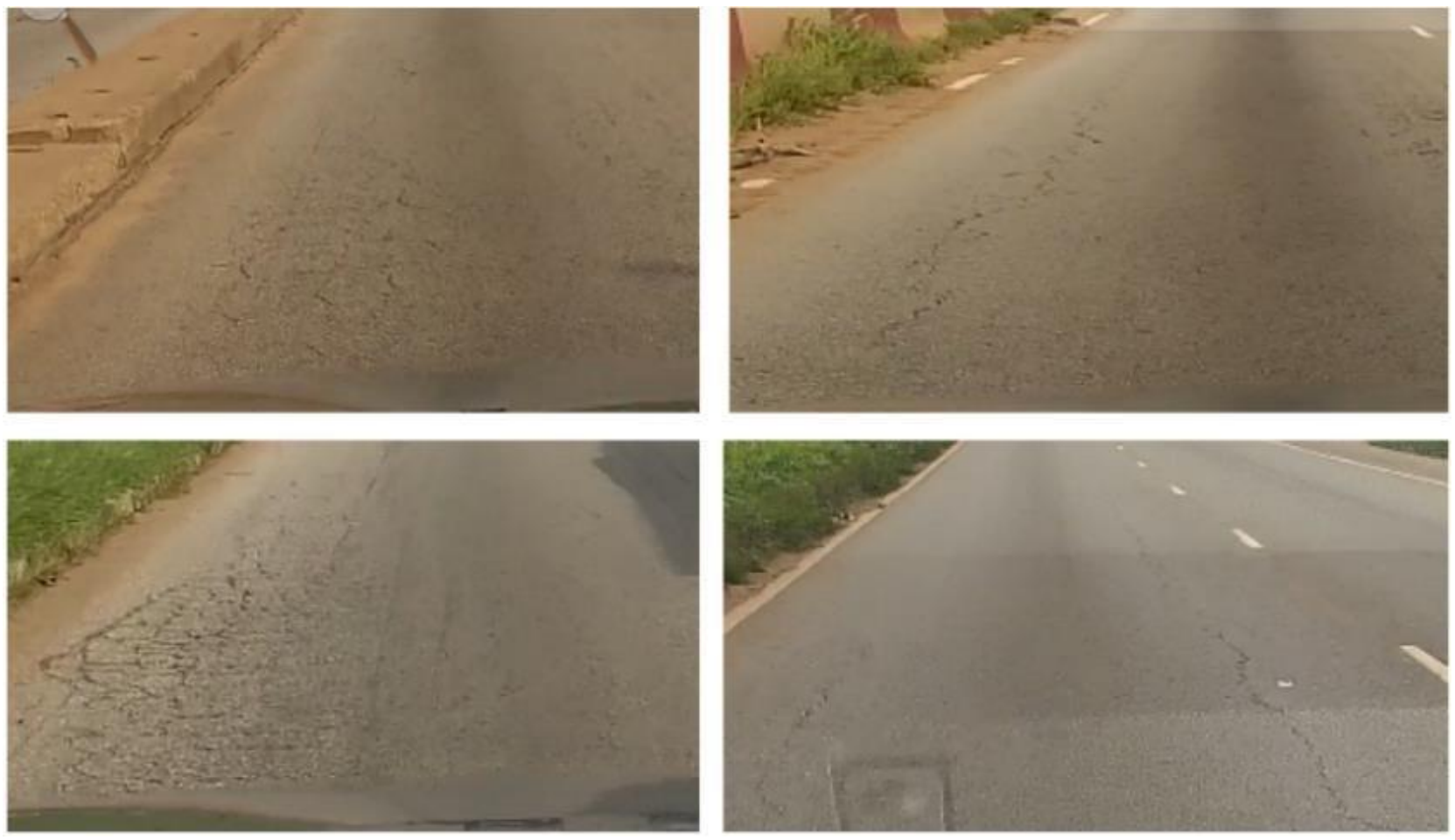

**Figure 5. Samples of low-quality data from the dashcam**

To enhance the dataset's diversity, pavement images from Google Street View were used. The platform's panoramic coverage of the target road networks was highly advantageous and though the image quality reduced when zooming to examine pavement surfaces closely, the platform proved to be very instrumental in providing information-rich images which contributed greatly to the variability of the dataset. Figure 6 shows samples of pavement images from Google Street view imagery.

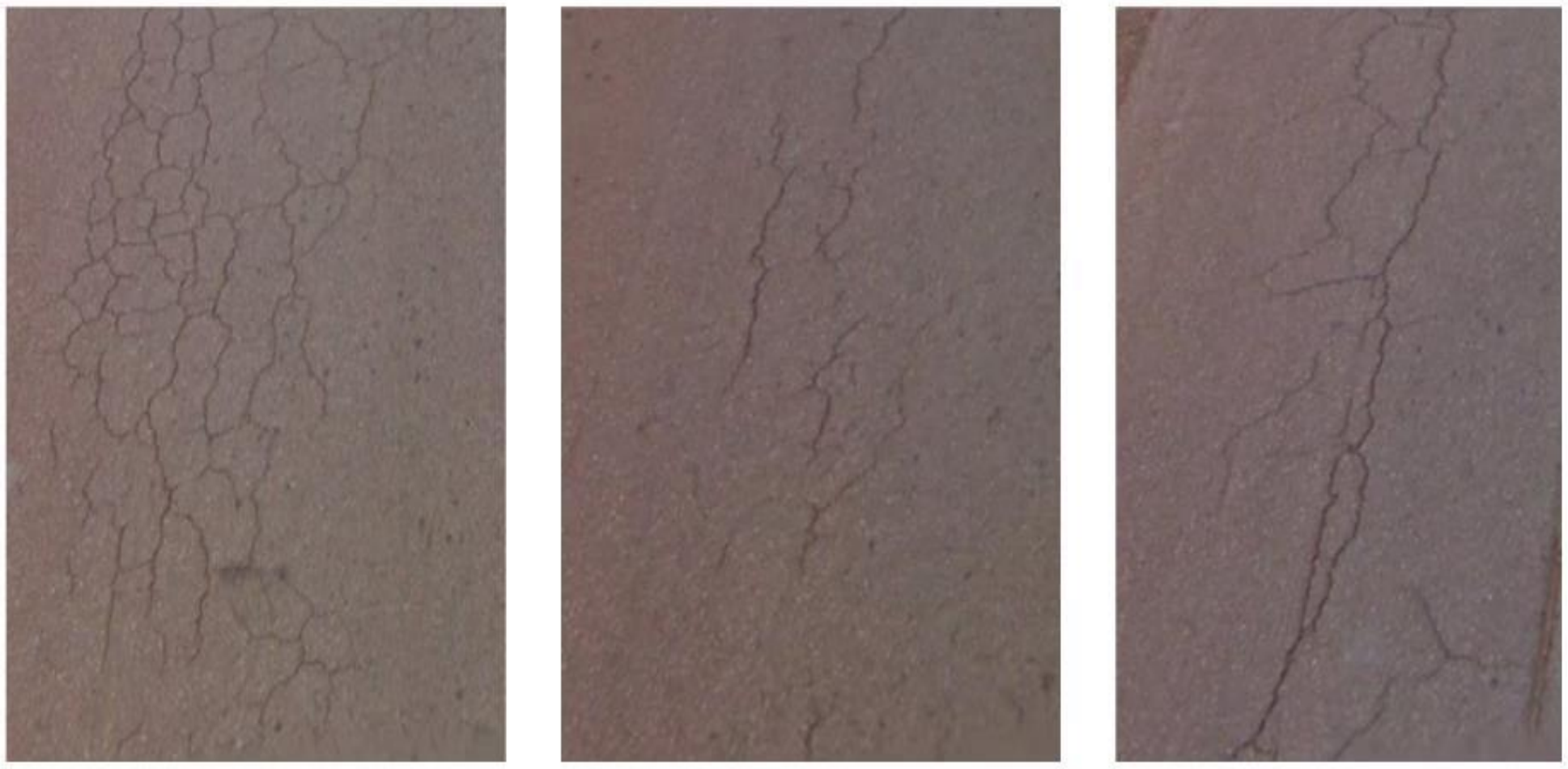

**Figure 6. Samples images from Google Street view imagery.**

The use of smartphones, despite being a slower method, offered greater control and higher image quality. This proved essential, as a good balance of low- and high-quality images were required to make the dataset robust. This approach also enabled capturing images from various angles and distances, ensuring a balanced representation of distress types and minimizing bias. Consequently, the quality of our training data was significantly enhanced, leading to improved accuracy in model predictions. Sample images of the data collected with smartphones are shown in Figure 7.

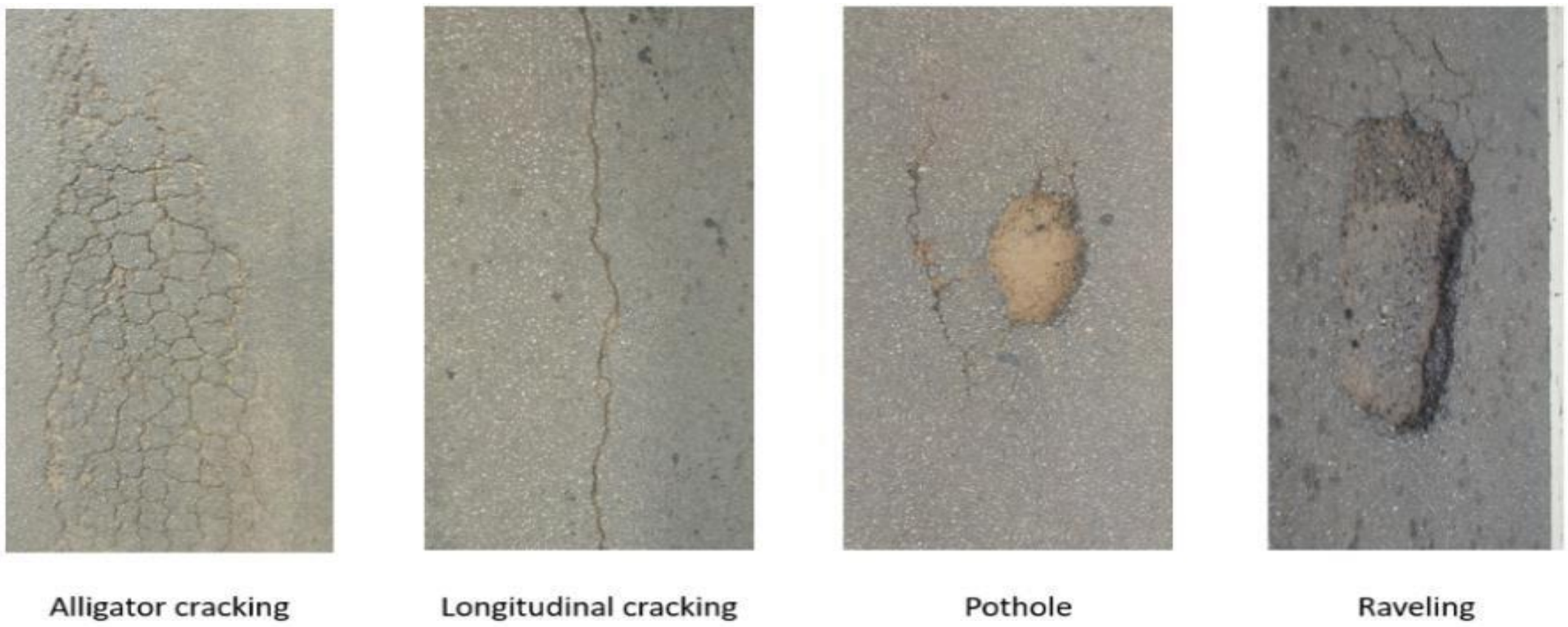

**Figure 7. Samples of images and their distress types collected with smartphones.**

To further diversify our dataset and enhance its generalizability, we incorporated images from online sources. These images, depicting diverse distress conditions, were specifically allocated for model testing, ensuring our model could handle variations not encountered in the manually collected data.

*Data preparation*

A multi-step data refinement process was implemented to ensure the quality and consistency of the dataset. First, all images underwent a quality check to identify and remove duplicates and irrelevant content. Duplicates were removed using computer programming, while irrelevant objects were manually cropped from images. To further ensure uniformity, all images were resized to a consistent dimension of 1000 x 2000 pixels, preserving crucial distress details while mitigating potential biases during model training. A clear and consistent naming convention, incorporating unique identifiers and information about the dominant distress type, was implemented to facilitate efficient data management and retrieval.

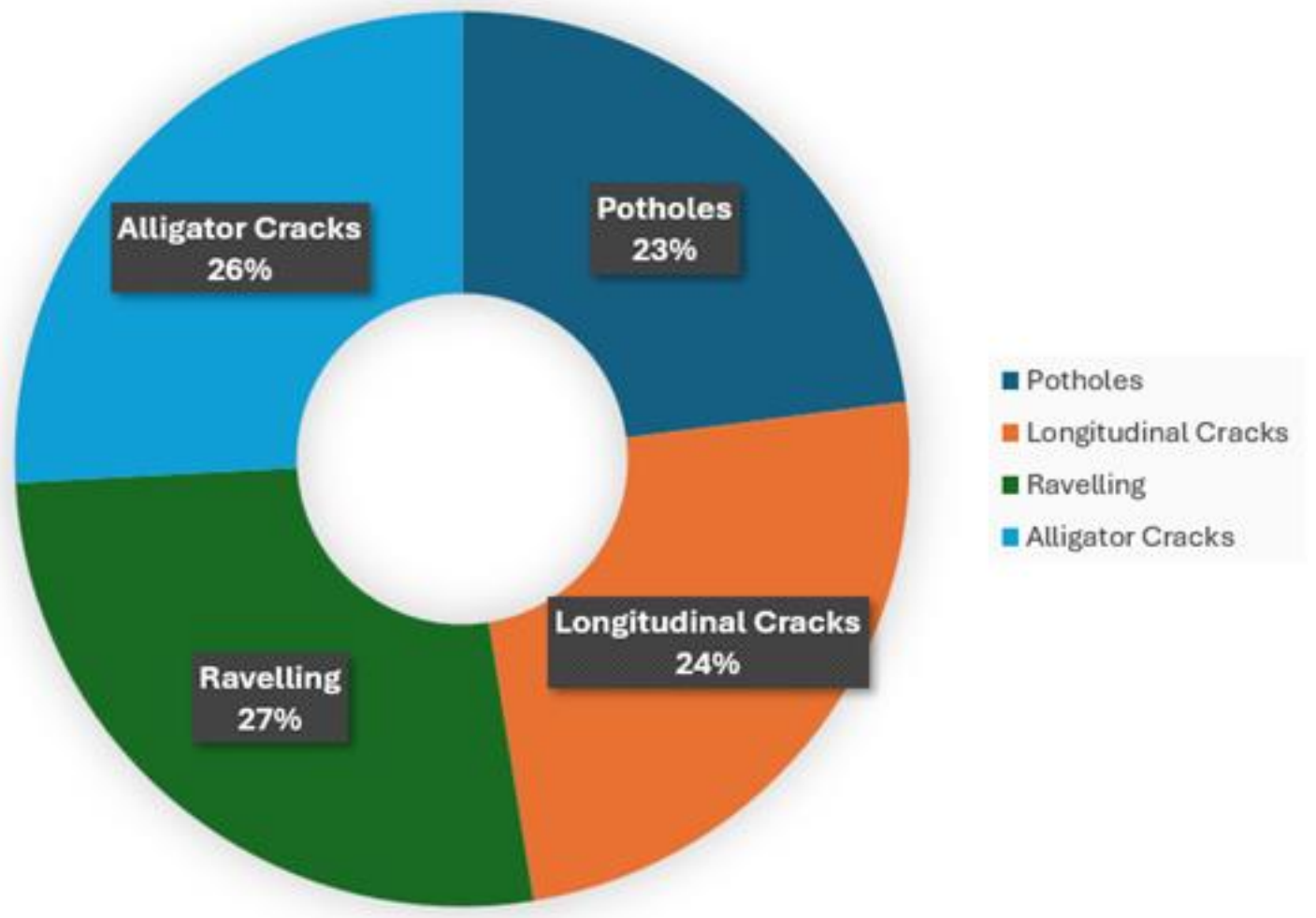

**Figure 8. A chart showing the statistics of the distress types after data collection**

Finally, using the Roboflow data annotation platform (*39*), each image was annotated, with bounding boxes drawn around identified distress types. These annotations provided essential spatial information for model training, enabling the a to learn and identify the visual characteristics of each distress category.

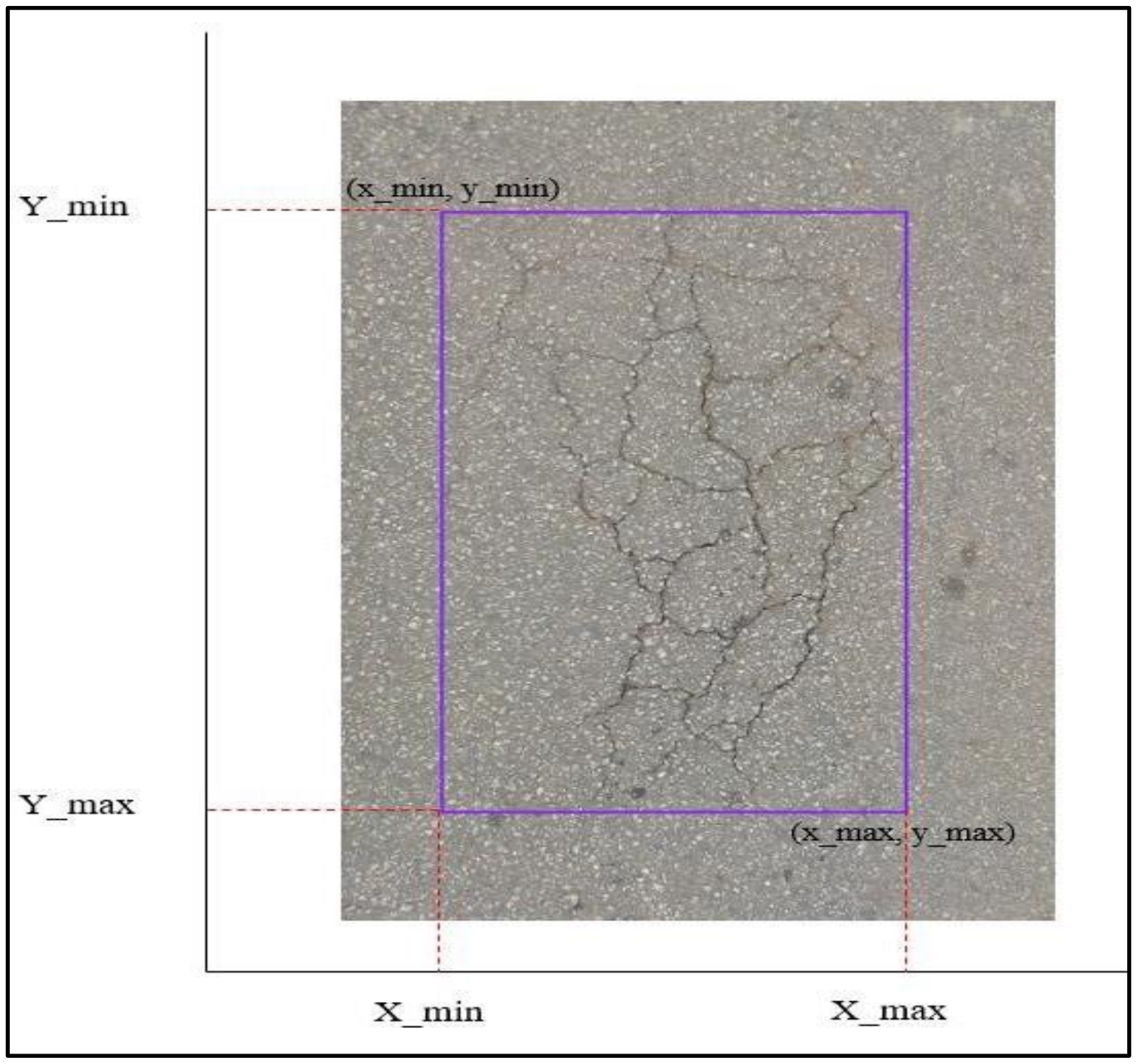

**Figure 9. Distress localization with bounding bow annotations**

The annotated dataset was split into training and validation sets, adhering to an 80:20 ratio. This division ensured that our models had sufficient data to learn effectively while reserving an independent subset for unbiased performance assessment. This rigorous preprocessing pipeline formed the bedrock of our model development process, ensuring that our pavement distress detection models are trained on a high-quality, consistent, and informative dataset.

## METHODOLOGY

Our approach to improving pavement distress detection in developing countries involves enhancing the YOLOv5 object detection model with advanced attention mechanisms. This section outlines the key components of our methodology, beginning with an overview of the Convolutional Block Attention Module (CBAM) and its integration into the YOLOv5 architecture.

### *Convolutional Block Attention Module (CBAM)*

Convolutional Block Attention Module (CBAM) is an attention mechanism that enables deep learning models focus on capturing the most informative features by computing channel and spatial attention maps [15]. Given an intermediate feature map $R^{C\times H\times W}$as input, CBAM computes two distinct attention maps along separate dimensions: a 1D channel attention map $M_C \in R^{C\times \mathbb{1}\times \mathbb{1}}$ and a 2D spatial attention map $M_S \in R^{\mathbb{1}\times H\times W}$. These attention maps are then used to generate the channel descriptor ($F'$) and final refined output ($F''$) as seen in Eq. 1 and Eq. 2:

$$F' = M_c(F) \otimes F \tag{1}$$

$$F'' = M_s(F') \otimes F' \tag{2}$$

The channel attention map $M_C$ is obtained through a series of operations. First, global average pooling (AvgPool(·)) and max pooling ($MaxPool(\cdot)$) operations are performed on the input feature map F. The resultant features from these operations ($F^c_{avg}$ and $F^c_{max}$) are projected through a shared lightweight multi-layer perceptron (MLP) with one hidden layer. The output feature maps from these resulting channel descriptors are then summed element-wise followed by a sigmoid activation function (σ). This enables the network to capture channel-wise dependencies by learning a weighted combination of the feature maps across channels. The channel attention map can be computed below:

$$\begin{aligned} M_C(F) &= \sigma\left(MLP\left(AvgPool(F)\right) + MLP\left(MaxPool(F)\right)\right) \\ &= \sigma\left(W_1\left(W_0\left(F^c_{avg}\right)\right) + W_1\left(W_0(F^c_{max})\right)\right), \end{aligned} \tag{3}$$

where $W_1$ and $W_0$ are the MLP weights.

The spatial attention map is derived by first applying global average pooling and global max pooling operations along the channel dimension of the input feature map F, resulting in two spatial descriptors ($F^s_{avg}$ and $F^s_{max}$ respectively). These spatial descriptors are then concatenated along the channel dimension. A convolutional layer ($Conv2D$) is applied to the concatenated feature, followed by a sigmoid activation function σ. The spatial attention map can be computed as below:

$$M_s(F) = \sigma\left(f^{7\times7}([AvgPool(F);\ MaxPool(F)])\right)$$

$$= \sigma\left(f^{7x7}\left([F^s_{avg}; F^s_{max}]\right)\right), \quad (4)$$

where $f^{7x7}$ performs a convolutional operation using a filter size of $7\ x\ 7$.

The channel attention module captures "what" features to focus on, while the spatial attention module identifies "where" those informative features are located within the input image. This helps the model focus on capturing the different distresses in the image and where those distresses are located. The structure of CBAM is illustrated in Fig. 10.

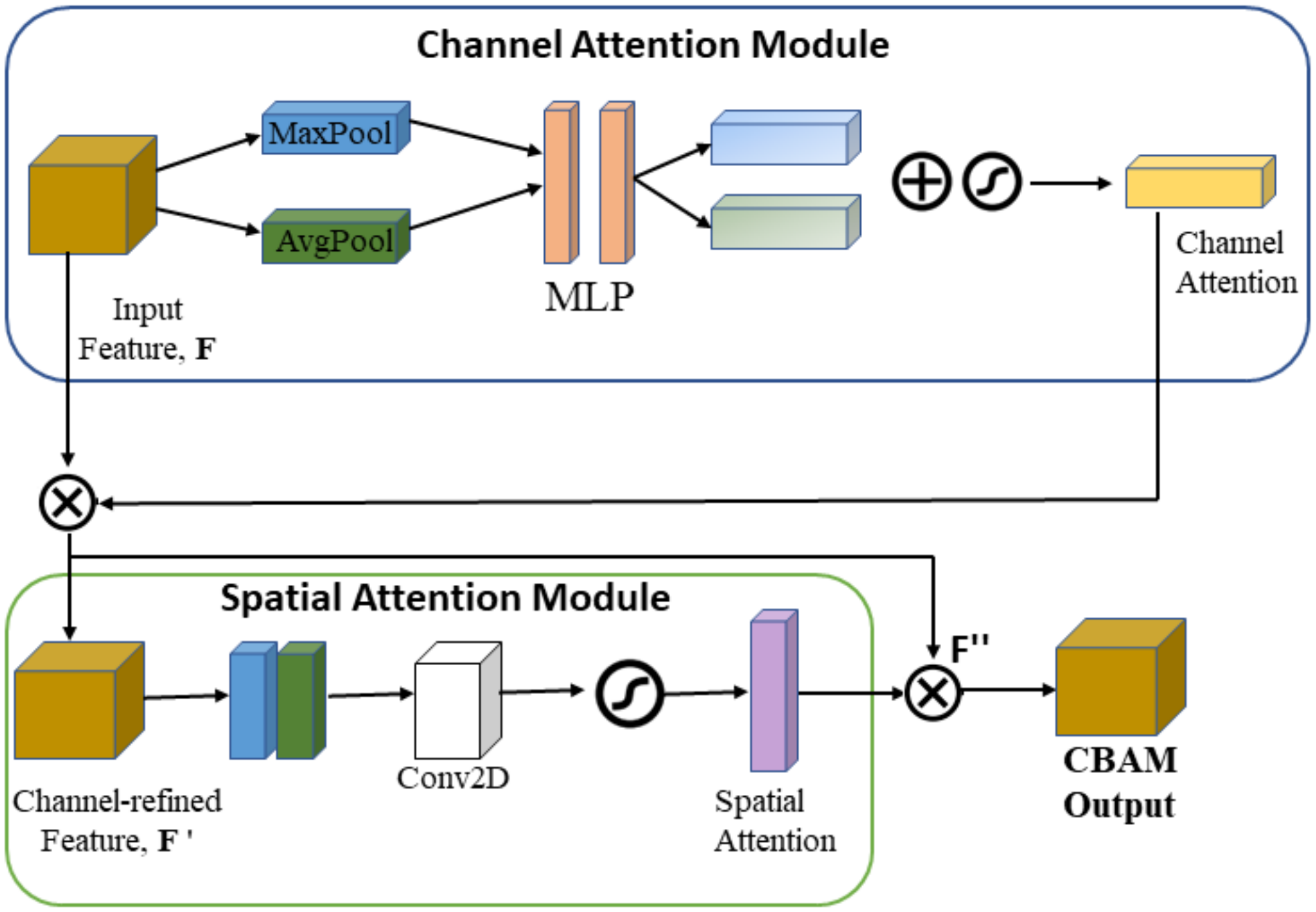

**Figure 10. CBAM architecture showing the Channel and Spatial Attention Module**

*Model architecture with an attention module*

We introduce the attention module (CBAM) into the YOLOv5 model to help the model capture the most informative parts of the images that contribute to the pavement distress detection. This modification is necessary to allow the model to focus on the subtle features and textures that characterize various types of pavement distresses. By incorporating CBAM, we enhance the model's ability to distinguish between different distress types, such as cracks and potholes, even in complex road surface conditions which is typical of developing countries.

### *2.3.1 YOLOv5 model with CBAM*

YOLOv5 model architecture comprises three primary components: a feature extraction backbone, a feature fusion neck, and a head. The backbone, responsible for extracting discriminative features, encompasses three integral modules: the Spatial Pyramid Pooling Fast (SPFF) layer, the Cross-Stage Partial Network (C3) blocks with 3 convolutions, and the Convolutional BatchNorm-SiLU (CBS) layers. The SPPF layer, preceded by a sequence of convolutional layers, leverages maximum pooling operations to enlarge the receptive field, thereby enhancing the nonlinear representational capacity of the network. The C3 block divides the feature maps into two parts and then merges them through a cross-stage hierarchy [19]. Furthermore, the CBS layers, which consist of a convolutional layer followed by batch normalization and the SiLU activation function, further process the features at different spatial scales. The next paragraph shows how the YOLOv5 architecture was modified.

For the YOLOv5 architecture, we replace the first convolutional layer in the C3 block with a CBAM. The CBAM output passes through Bottleneck layers and concatenates with two additional convolutional layers' output as shown in Fig. 11. By replacing the first convolutional layer of the C3 block with CBAM, the model can adaptively refine the feature maps at an early stage. This process emphasizes the most informative channels and spatial regions before further processing through the remaining layers of the C3 block and the subsequent bottleneck layers. Furthermore, the concatenation of the resulting CBAM output with the output of the two additional convolutional layers within the C3 block, ensures that the attention-modulated features are combined with the traditional convolutional features. This concatenation allows the model to leverage both the attention-based feature refinement and the conventional convolutional feature extraction, potentially leading to a more robust and discriminative representation of the input data. The overall network structure of the YOLOv5 with an attention module is shown in Fig 12.

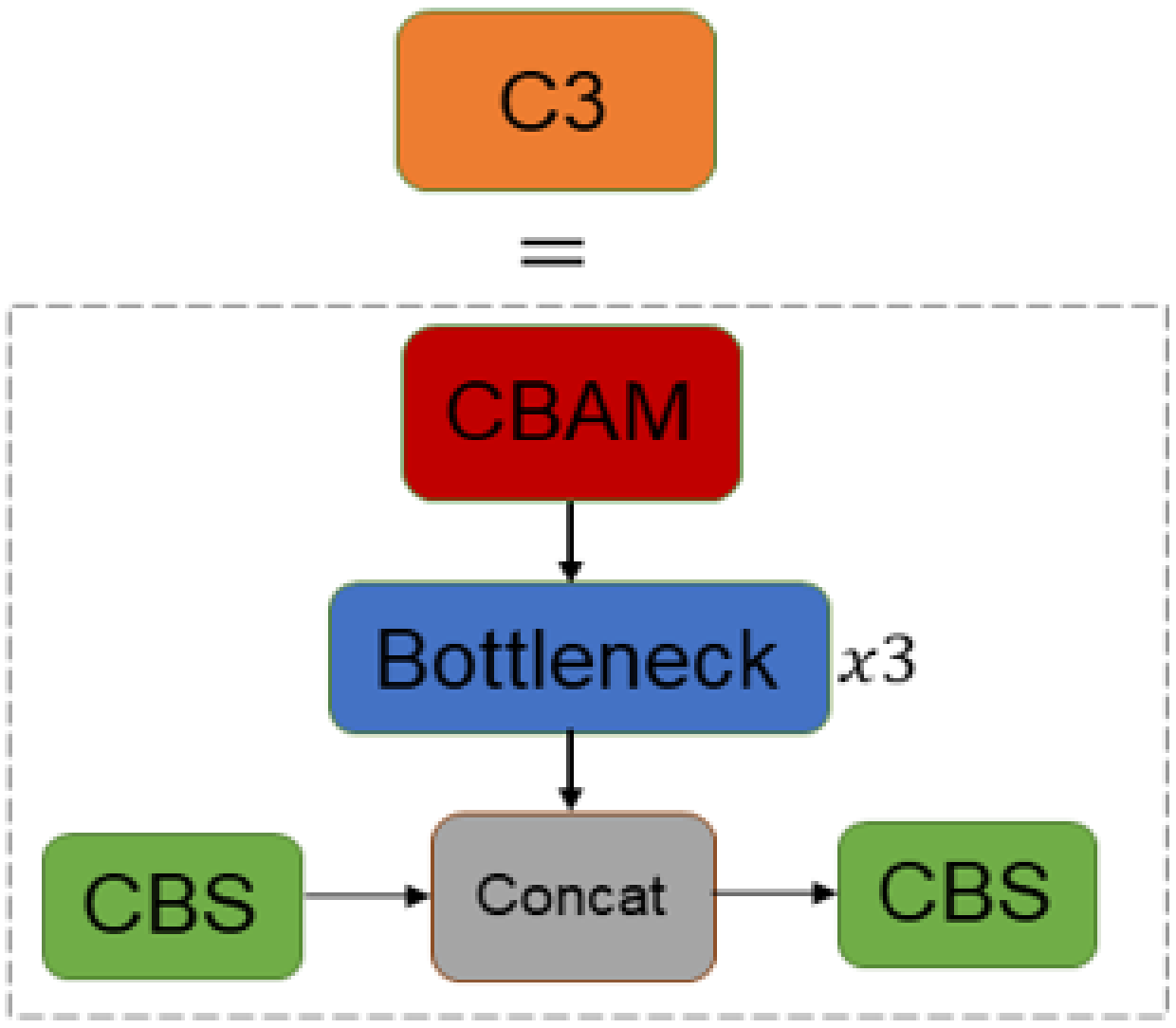

**Figure 11. Modified C3 block in the YOLOv5 architecture**

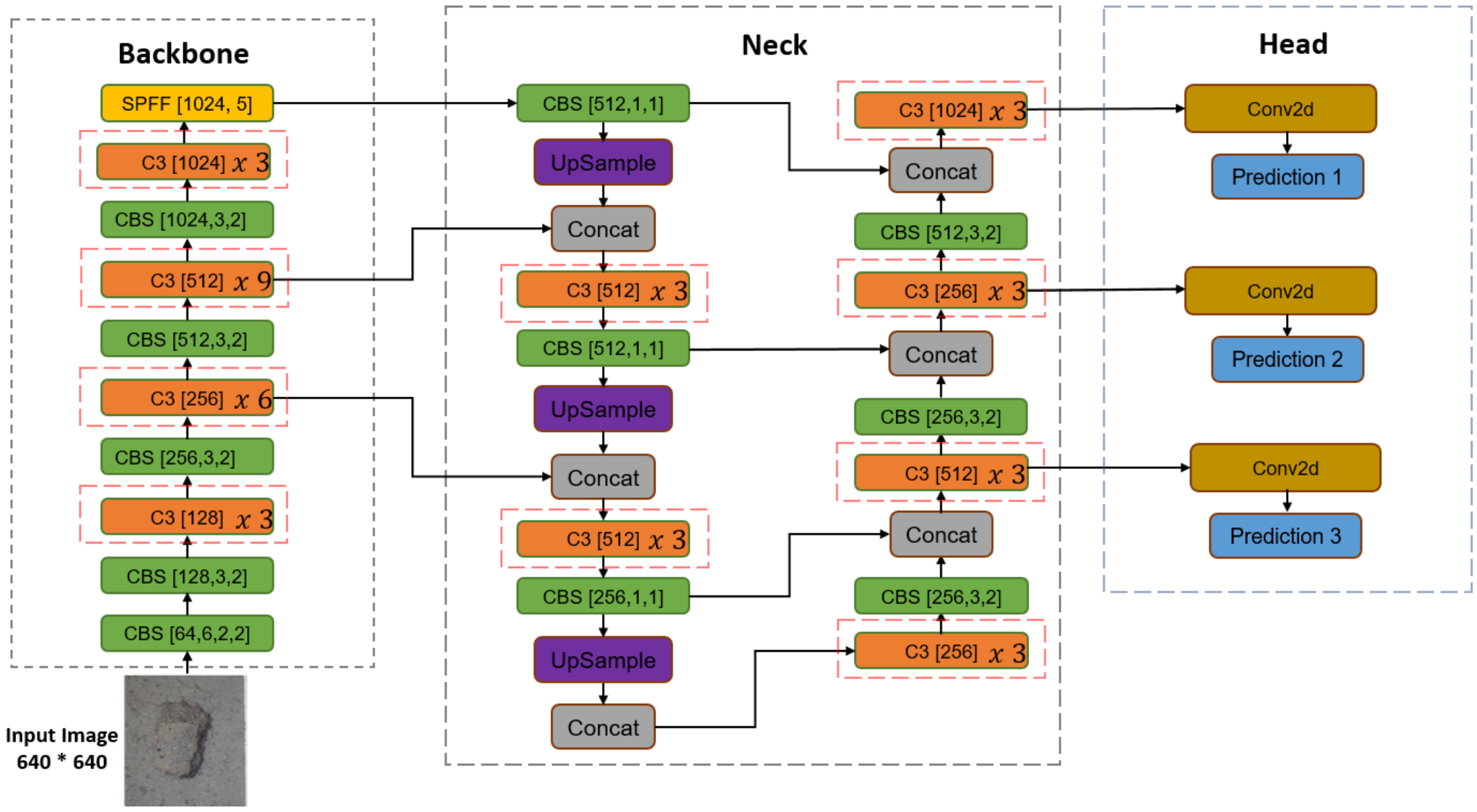

**Figure 12. Overall YOLOv5 architecture with modified C3 blocks**. The dotted red rectangular boxes show the areas where the architecture was modified.

## Evaluation Metrics

In evaluating the performance of the pavement distress detection model, we utilized three critical metrics: Precision, Recall, and Mean Average Precision (mAP). The details of each of the metrics has been illustrated below.

### *Precision*

Precision is a metric that measures the accuracy of the model's positive predictions. It is calculated as the ratio of true positives to the total number of predicted positives:

$$Precision = \frac{TP}{TP + FP} \tag{5}$$

where $TP$ is the number of true positives (correct detections) and $FP$ is the number of false positives (incorrect detections). Precision evaluates the proportion of correctly identified distress instances among all instances labeled as distress by the model.

### *Recall*

Recall, also known as sensitivity or true positive rate, measures the model's ability to identify all relevant instances of distress. It is calculated as:

$$Recall = \frac{TP}{TP + FN} \tag{6}$$

where $FN$ is the number of false negatives (missed detections). Recall highlights the model's effectiveness in capturing all true instances of distress, thereby indicating how well the model performs in minimizing missed detections.

*Mean Average Precision (mAP)*

Mean Average Precision (mAP) is a comprehensive metric that assesses the overall accuracy of the model by averaging the precision across various recall levels and IoU (Intersection over Union) thresholds. For a predicted bounding box $B_P$ and a ground truth bounding box $B_{gt}$, the IoU is computed as:

$$IoU = \frac{|B_P \cap B_{gt}|}{|B_P \cup B_{gt}|} \tag{7}$$

Precision and Recall are calculated at various IoU thresholds $t$ as follows:

$$Precision(t) = \frac{TP(t)}{TP(t) + FP(t)} \tag{8}$$

$$Recall(t) = \frac{TP(t)}{TP(t) + FN(t)} \tag{9}$$

The Average Precision (AP) for a specific class is then computed as:

$$AP = \int_0^1 Precision(r)\, dr \tag{10}$$

This integral is then approximated by summing precision values at different recall levels. The mAP is the mean of the AP values across all classes:

$$mAP = \frac{1}{C}\sum_{c=1}^{C} AP_C \tag{11}$$

where $C$ is the total number of classes.

**RESULTS AND DISCUSSIONS**

Figure 13 illustrates the training and validation losses over 200 epochs for box loss, object loss, and classification loss. The training losses show a steady decline, with box loss decreasing from around 0.10 to 0.04, object loss from 0.07 to 0.05, and classification loss from 0.04 to 0.01. Similarly, the validation losses exhibit a comparable trend, with box loss reducing from 0.11 to 0.05, object loss from 0.05 to 0.025, and classification loss from 0.04 to 0.02. These trends suggest effective learning and good generalization, as the validation losses closely follow the training losses, indicating minimal overfitting.

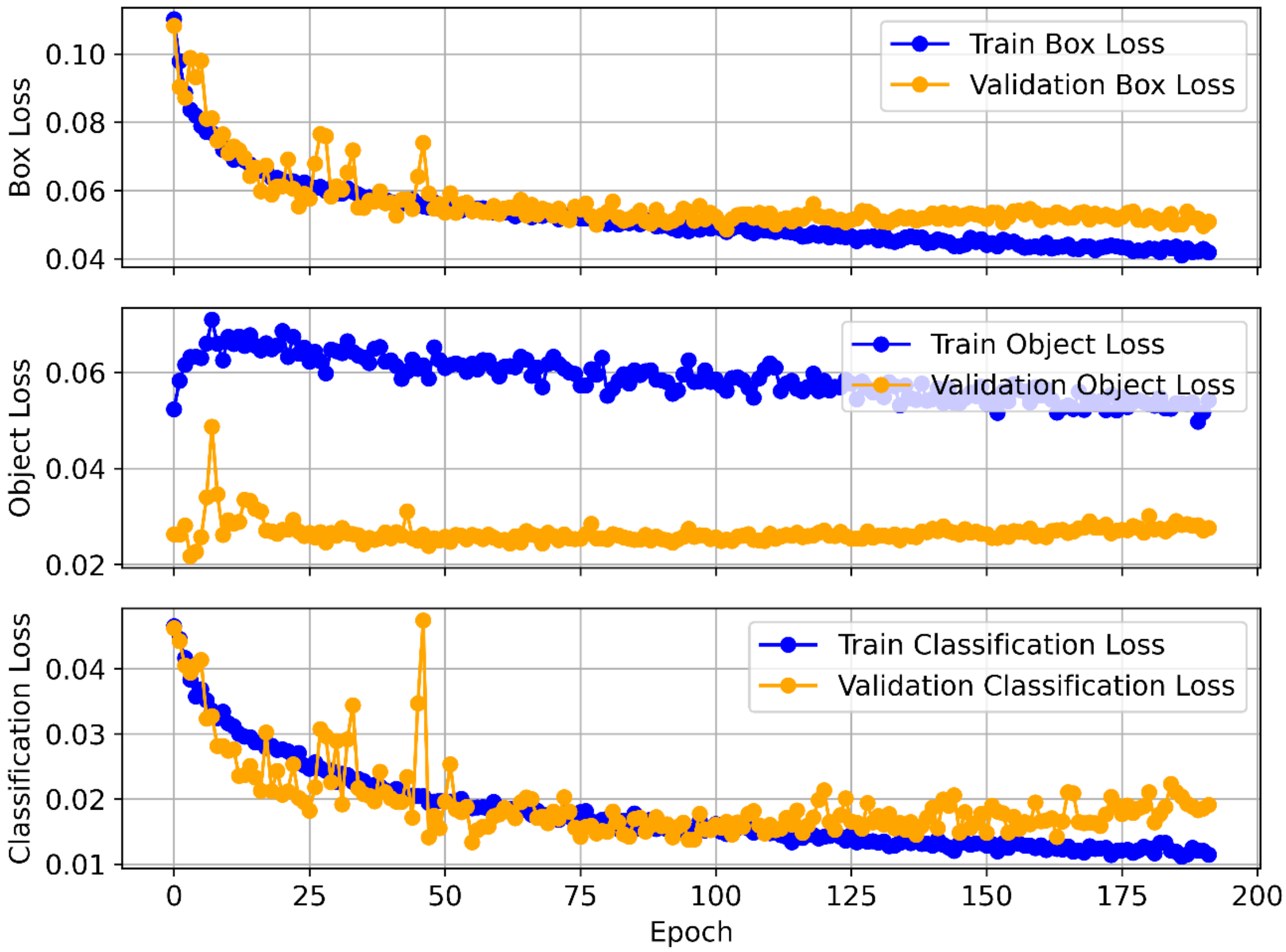

**Figure 13. Training and validation losses over 200 epochs for box loss, object loss, and classification loss**

Additionally, the analysis reveals key trends in model performance. These trends are illustrated in Figure 14. At lower confidence levels, recall is high, indicating that the model detects most true positives but includes many false positives. As confidence increases, recall decreases, reflecting more conservative predictions with fewer false positives. Precision, on the other hand, starts low but improves with higher confidence levels, showing a reduction in false positives as the model's certainty increases.

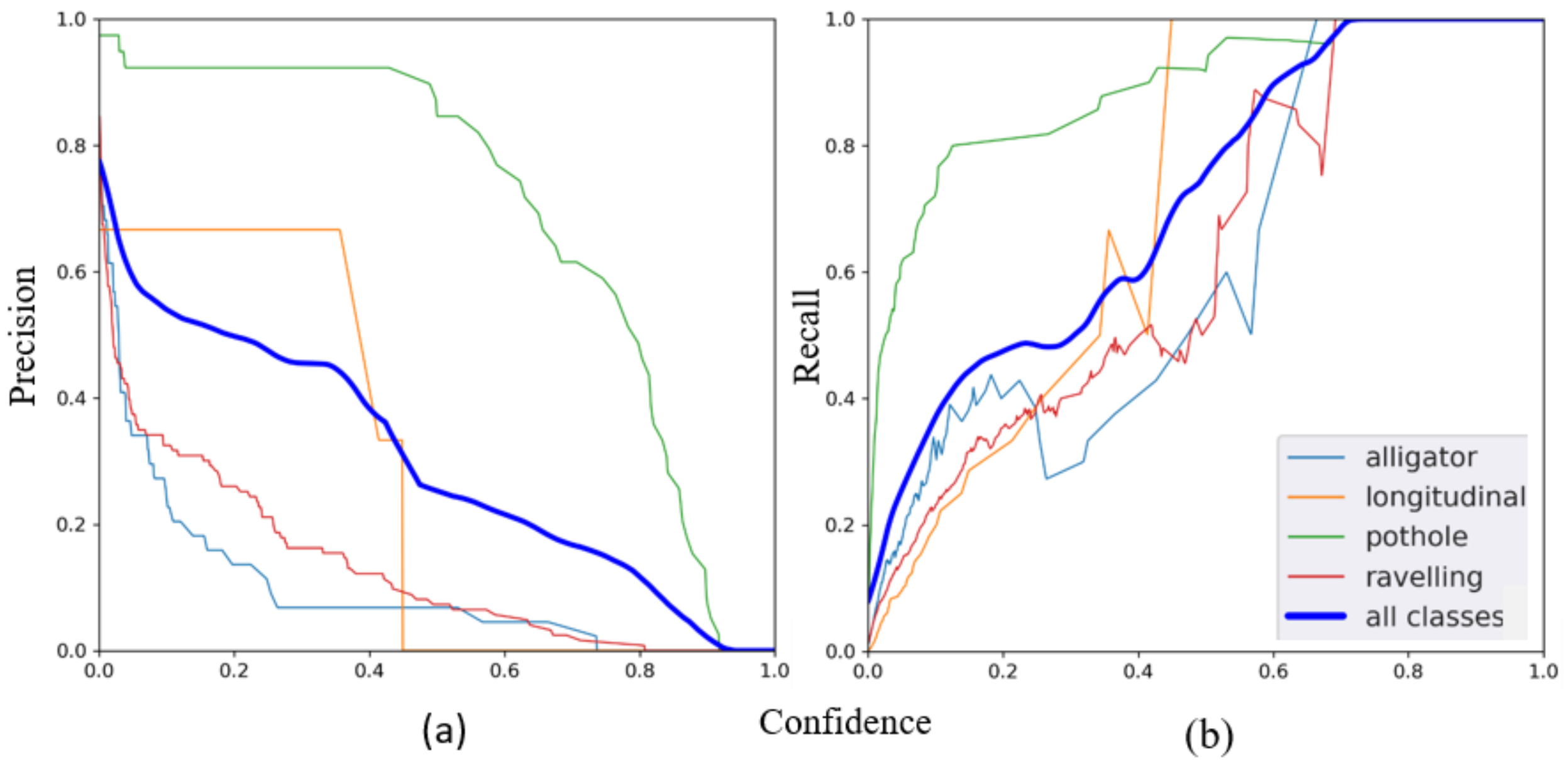

**Figure 14. Illustration of the trade-off between recall and precision across different confidence levels**. (a) Precision-Confidence Curve (b) Recall-Confidence Curve

*Qualitative Results*

The qualitative results demonstrate the model's ability to detect and classify various pavement distresses, including potholes, longitudinal cracks, alligator cracks, and ravelling. The provided images highlight the model's precision in localizing these defects, with confidence scores ranging from 0.46 to 0.93. For instance, the model accurately detects potholes with confidence scores of 0.89 and 0.93, while ravelling is identified with scores such as 0.88 and 0.85. The detection of multiple distress types within a single image, such as the co-occurrence of longitudinal cracks and potholes, showcases the model's capability for multi-class detection. The precise bounding boxes generated indicate the model's effectiveness in capturing the spatial extent of the distresses, crucial for subsequent analysis and maintenance planning. These qualitative results confirm the model's robust performance across various pavement conditions, making it a reliable tool for automated pavement distress detection and classification. The results have been illustrated in Figure 15.

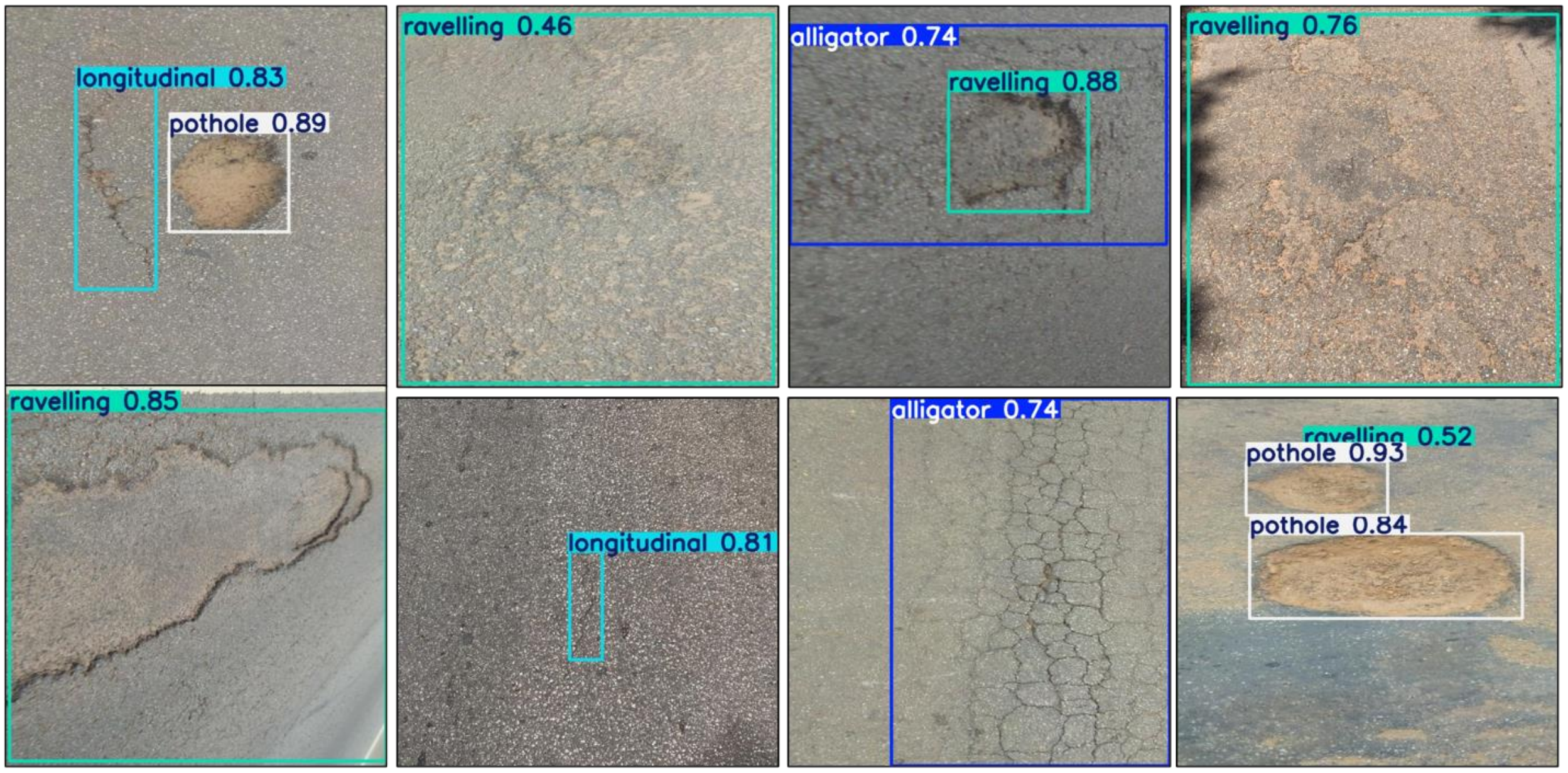

**Figure 15. Examples of accurate distress detections.** The model successfully identifies and classifies distresses such as longitudinal cracks, potholes, alligator cracks, and ravelling with high confidence scores

While the model generally performs well, there are instances where it struggles with accurate detection and classification, leading to errors. For example, in Figure 16(a), the model incorrectly classified a distress as raveling instead of identifying it as a pothole. In Figure 16(b), the model failed to detect the presence of raveling and falsely identified alligator cracking where none was present. Additionally, Figure 16(c) shows multiple false detections, where the model erroneously detected several instances of ravelling that do not align with the actual pavement condition. Finally, in Figure 16(d), the model misclassified part of the image, detecting only ravelling and overlooking the alligator cracking present, while also incorrectly identifying a portion of the image as ravelling.

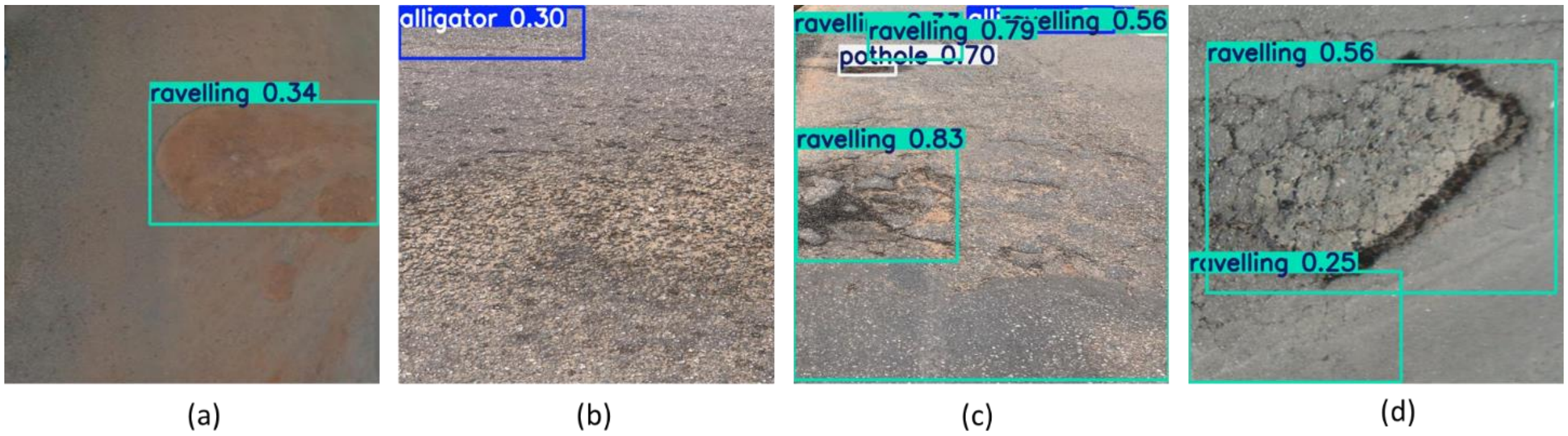

**Figure 16. Examples of misclassifications in pavement distress detection.** (a) Misclassification of a pothole as ravelling, (b) Failure to detect ravelling and false detection of alligator cracking, (c) Multiple false detections of ravelling, and (d) Misclassification of alligator cracking as ravelling and detection of non-existent distress. These errors highlight challenges in accurately detecting and classifying pavement distresses in varied conditions commonly found in developing countries.

These inaccuracies can be attributed to the challenging conditions often encountered in pavement surfaces within developing countries. The subtle and complex nature of certain distress patterns, compounded by variations in pavement texture due to differing construction materials and techniques, poses significant challenges for the model. Additionally, inconsistent lighting conditions and shadows, which are common in outdoor settings in developing regions, further complicate accurate detection. The inherent difficulty of distinguishing between similar distress types, such as ravelling and potholes, is exacerbated by the mixed defect types often found in these environments. Moreover, the model's occasional misclassifications highlight the limitations of existing datasets, which may not fully capture the unique characteristics of pavement distresses in these regions.

*Grad-CAM*

Grad-CAM (Gradient-weighted Class Activation Mapping) is a technique used to visualize where the model is focusing when making predictions. By generating attention maps, Grad-CAM highlights the regions in the input image that contribute most to the model's predictions. This is particularly useful in understanding the model's behavior and ensuring that it is focusing on relevant features of the pavement distresses, such as cracks, potholes, or areas of surface deterioration. Figure 17 shows images with their corresponding detections and attention maps illustrating areas the model focused on the distress detection. The visualization provides insights into the model's interpretability, allowing us to verify whether the model is accurately identifying the distress locations or being misled by irrelevant features. By analyzing these attention maps, we can better understand the strengths and weaknesses of the model's detection process.

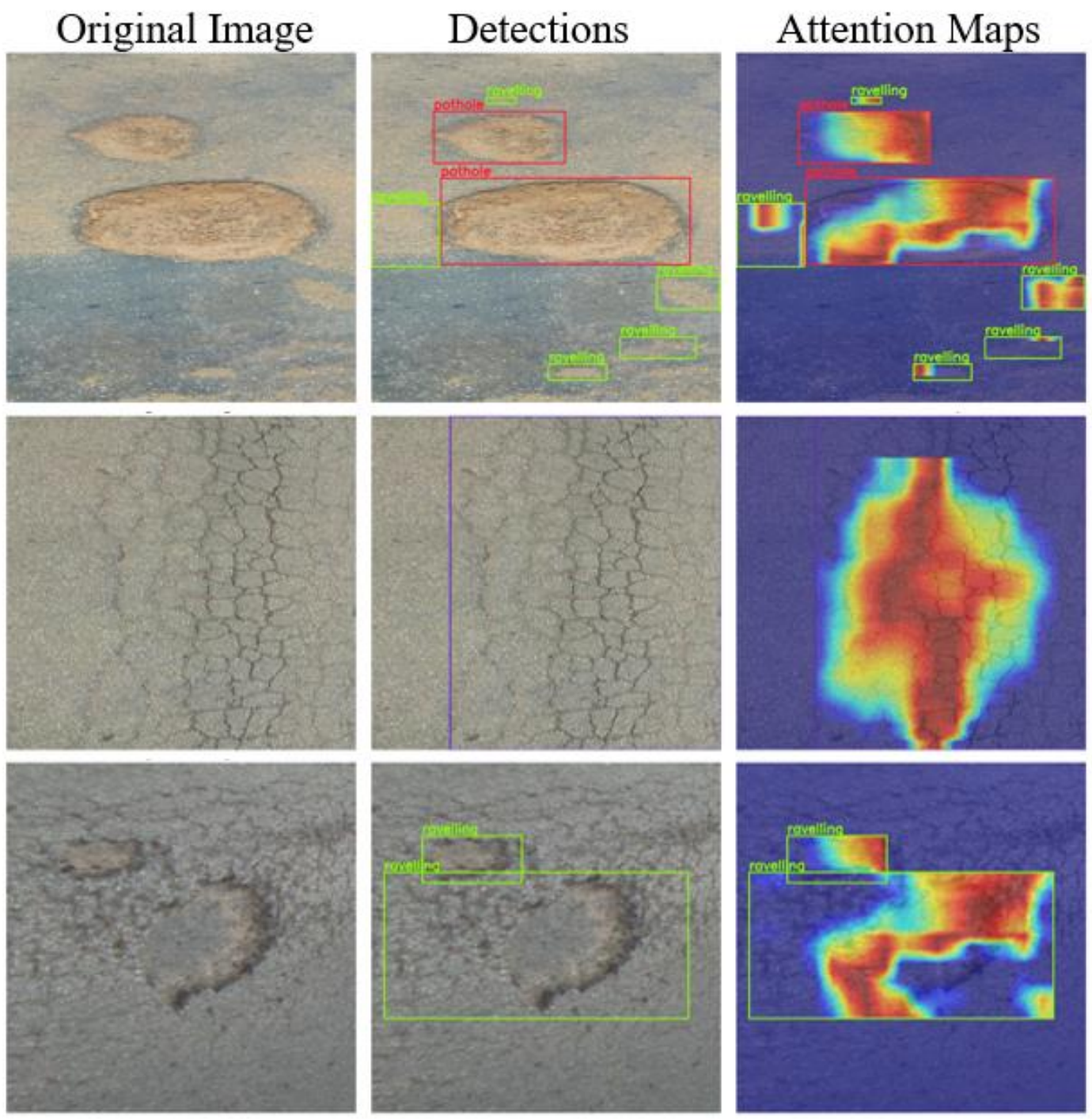

**Figure 17. Grad-CAM visualizations demonstrating the model's focus areas during detection**. The original images are compared with the model's detections and their corresponding

attention maps, which highlight the regions the model considered most influential in identifying pavement distresses such as potholes, alligator cracks, and ravelling.

*Web-Based Platform for Real-Time Detection and Classification*

In order to address the critical need for efficient infrastructure management in developing countries, we designed a web-based platform optimized for the real-time detection and classification of pavement distresses. This platform is versatile, offering both image-based and video-based inference capabilities, which cater to the varied data collection methods commonly used in these regions.

Image-Based Inference

The image-based inference system, illustrated in Figure 18, enables users to upload individual images of asphalt pavement surfaces. Once uploaded, the model processes these images to identify and classify different types of pavement distresses, such as raveling, potholes, and cracks. The system then displays the results with bounding boxes and confidence scores, which indicate the model's certainty in each prediction. This functionality is highly practical for infrastructure managers in developing countries, where the availability of resources for continuous monitoring might be limited. By using smartphones or basic digital cameras, field workers can easily capture images of road surfaces, which can then be analyzed by the platform. This allows for rapid assessment of road conditions without requiring sophisticated equipment or extensive training

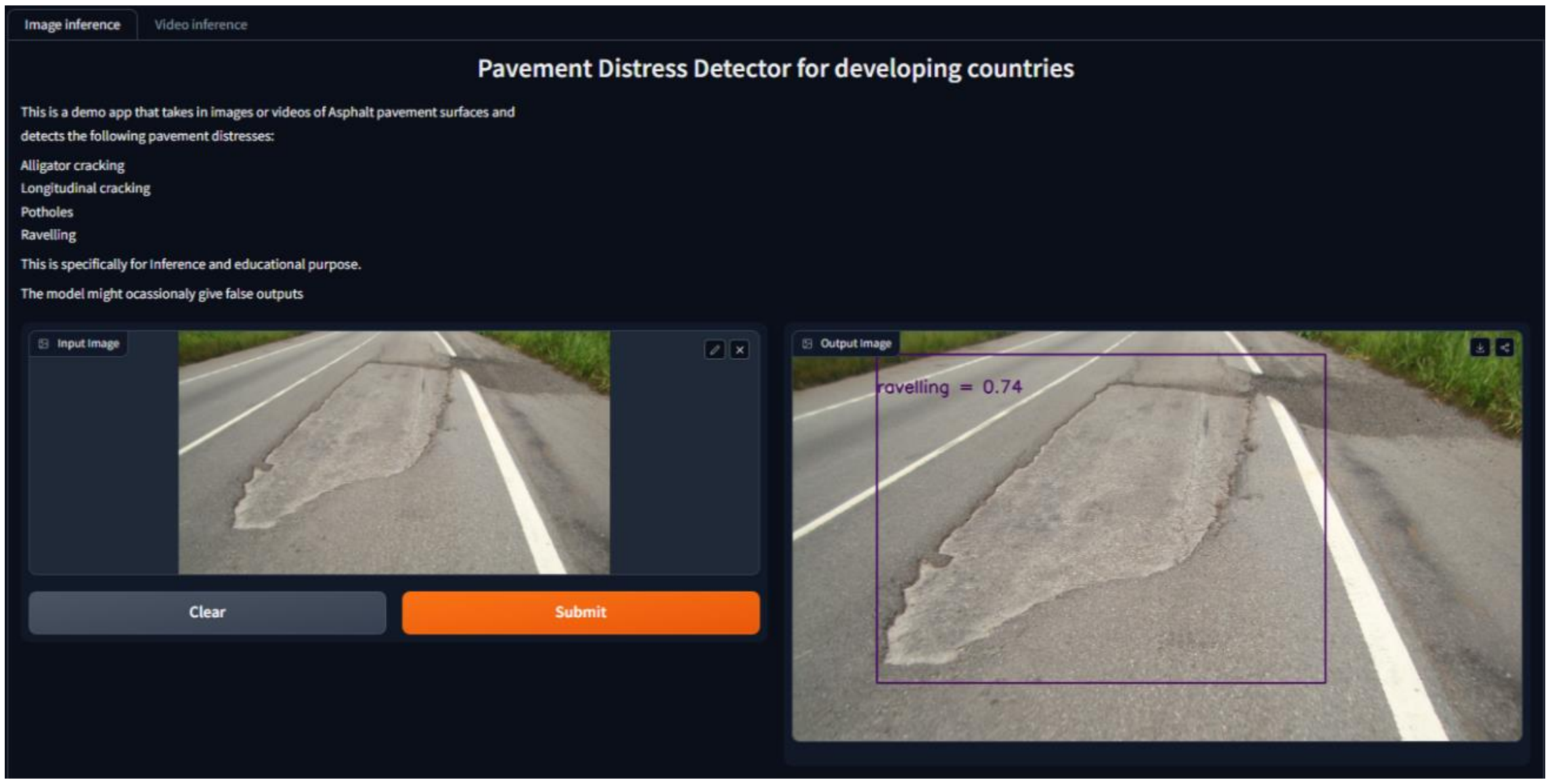

**Figure 18. Image-based inference system detecting ravelling on an asphalt pavement with a confidence score of 0.74.**

Video-Based Inference

The video-based inference system, shown in Figure 19, extends the platform's capabilities to continuous video streams, making it suitable for real-time monitoring of road networks. This system processes each frame of the video to detect and classify pavement distresses, such as alligator cracking, and provides real-time feedback by displaying the detected distresses with bounding boxes on the video feed. In practical terms, this feature is invaluable for road maintenance operations in developing countries, where road conditions can deteriorate rapidly due to harsh weather, heavy traffic, and limited maintenance resources. By equipping vehicles or drones with cameras, infrastructure managers can continuously monitor the condition of roads as they are being used. This real-time data collection allows for immediate identification of areas needing repair, which is crucial for maintaining road safety and prolonging the lifespan of the infrastructure.

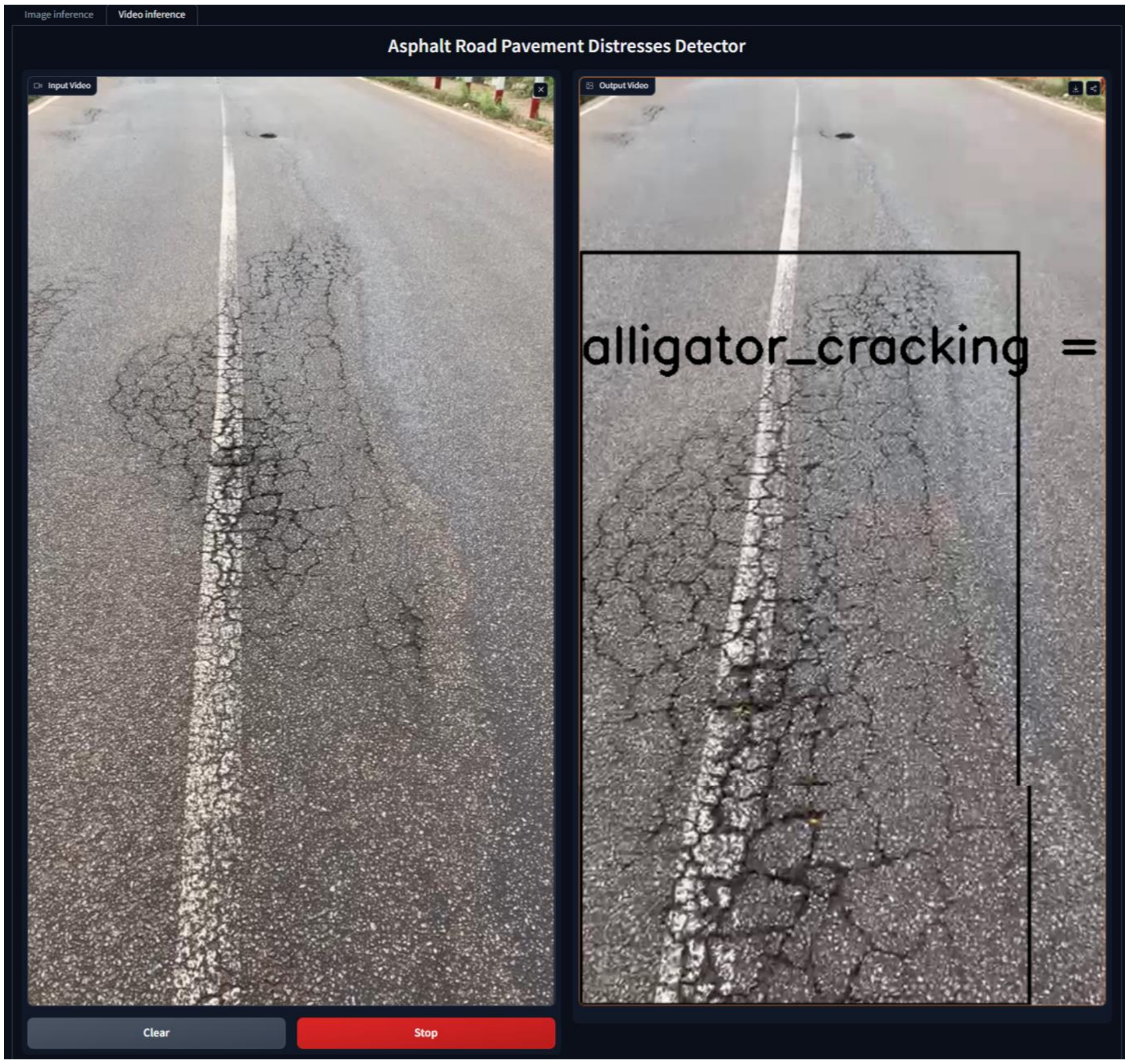

**Figure 19. Video-based inference system detecting alligator cracking on an asphalt pavement.** The system processes video feeds in real-time, identifying and classifying pavement distresses as they occur.

## CONCLUSIONS

This study addressed the critical challenge of pavement distress detection in developing countries by introducing a novel deep learning approach that combines YOLOv5 with Convolutional Block Attention Module (CBAM). Our YOLOv5-based model, enhanced with CBAM, showed remarkable efficacy in focusing on subtle distress features in complex environments, achieving high confidence scores ranging from 0.46 to 0.93 in detecting potholes, longitudinal cracks, alligator cracks, and raveling. While some misclassifications in complex scenarios were observed, these instances provide valuable insights into the unique challenges of pavement assessment in developing countries and offer direction for future improvements. In addition, we create and make publicly available pavement distresses dataset comprising of high-quality pavement images in developing countries. This effort addresses a significant gap in existing research in pavement condition assessment in developing countries.

The development of a web-based application for real-time distress detection represents a significant step towards practical implementation, offering a powerful tool for infrastructure management in resource-constrained environments. This research has far-reaching implications for improving road safety, optimizing maintenance strategies, and contributing to sustainable transportation infrastructure development in developing regions.

Future work should focus on expanding the dataset to encompass a wider range of pavement conditions and distress types, further refining the model's performance in challenging scenarios, and integrating the system with existing infrastructure management practices. Additionally, exploring the potential of this technology for predictive maintenance could further enhance its value for long-term infrastructure planning.

## AUTHOR CONTRIBUTIONS

The authors confirm contribution to the paper as follows: study conception and design: Blessing Agyei Kyem, Kenneth Adomako Tutu; data preparation and collection: Blessing Agyei Kyem, Kenneth Adomako Tutu, Eugene Kofi Okrah Denteh; analysis and interpretation of results: Blessing Agyei Kyem, Armstrong Aboah; draft manuscript preparation: Blessing Agyei Kyem, Joshua Kofi Asamoah. All authors reviewed the results and approved the final version of the manuscript.